\documentclass[letterpaper, 10 pt, conference]{ieeeconf}

\IEEEoverridecommandlockouts                             

\usepackage{graphicx}
\usepackage{xcolor}
\usepackage{booktabs}
\usepackage[width=.75\textwidth]{caption}

\usepackage[backend=biber,
style=numeric,
bibstyle=ieee,
citestyle=numeric-comp,
sorting=none, 
maxbibnames=99]{biblatex}
\title{\LARGE \bf
An Evolutionary Framework for Connect-4 as Test-Bed for Comparison of Advanced Minimax, Q-Learning and MCTS
}

\author{Henry Taylor and Leonardo Stella
\thanks{H. Taylor and L. Stella are with the School of Computer Science, College of Engineering and Physical Sciences, University of Birmingham, Birmingham, B15 2TT, U.K. {\tt\small(email: hxt138@alumni.bham.ac.uk, l.stella@bham.ac.uk)}.
}
}

\begin{document}

\maketitle
\thispagestyle{empty}
\pagestyle{empty}

\begin{abstract}
A major challenge in decision making domains with large state spaces is to effectively select actions which maximize utility. In recent years, approaches such as reinforcement learning (RL) and search algorithms have been successful to tackle this issue, despite their differences. RL defines a learning framework that an agent explores and interacts with. Search algorithms provide a formalism to search for a solution. However, it is often difficult to evaluate the performances of such approaches in a practical way. Motivated by this problem, we focus on one game domain, i.e., Connect-4, and develop a novel evolutionary framework to evaluate three classes of algorithms: RL, Minimax and Monte Carlo tree search (MCTS). The contribution of this paper is threefold: i) we implement advanced versions of these algorithms and provide a systematic comparison with their standard counterpart, ii) we develop a novel evaluation framework, which we call the \textit{Evolutionary Tournament}, and iii) we conduct an extensive evaluation of the relative performance of each algorithm to compare our findings. We evaluate different metrics and show that MCTS achieves the best results in terms of win percentage, whereas Minimax and Q-Learning are ranked in second and third place, respectively, although the latter is shown to be the fastest to make a decision.
\end{abstract}

\section{INTRODUCTION}

Recent successes of game artificial intelligence (AI) have sparked increasing interest in the research community. Examples include Deep Blue, beating the ruling World Chess Champion Grandmaster Gary Kasparov in 1997, and Deepmind AlphaGo \cite{Silver2016}, beating European Go Champion Fan Hui and Go Master Lee Sedol in a five-game match. These recent achievements have led to a focus on General Game Playing (GGP) algorithms where single agents can learn multiple games to super-human level. The problem that arises when comparing these different algorithms is the lack of consistency in methodologies.

The aim of this paper is to provide a thorough analysis of a range of algorithms including Q-learning, Minimax and Monte Carlo Tree Search (MCTS), as well as their more advanced counterparts. To address the issue of inconsistency in the analysis, we carry out the comparison of these approaches across a single game domain, i.e., Connect-4. Connect-4 is a two-player zero-sum perfect information game played on a vertical board of size $6 \times 7$ - consisting of over 4.5 trillion game states. The outcome of the game for either player can be win, lose or draw; these conditions are referred to as \textit{terminal states} and a win is awarded if four pieces are connected horizontally, vertically or diagonally \cite{kang2019, Schneider2002}. Connect-4 was chosen as a test-bed for comparison in line with the large body of works that constitute the literature review of this paper as in the following.

Connect-4 was first independently solved by Allen and Allis in 1988 via two different approaches \cite{VANDENHERIK2002}. Allen used a Brute-Force Depth-First search while Allis built a program called \textit{VICTOR} which used knowledge-based strategic rules based on the Chess concept of Zugzwang \cite{VANDENHERIK2002}. More recently, the authors in \cite{Wang2019} applied Q-Learning with Epsilon-Greedy policy to small board sizes for Connect-4 and other games. In \cite{kang2019}, different Minimax payoff functions in Connect-4 were investigated, resulting in complex heuristics being advantageous at deeper Minimax searches. In \cite{Nasa2018}, Minimax with Alpha-Beta pruning cuts was found to be an improvement over the base Minimax algorithm for Connect-4. The authors in  \cite{Scheiermann_2022} applied a Reinforcement Learning (RL) agent with a MCTS wrapper to several games including Connect-4. They showed that a base MCTS agent (with random play-outs and 10,000 iterations) had a win rate of $\approx$100\% against the RL agent, $\approx$25\% against a near perfect Minimax Alpha-Beta agent, and $\approx$1\% against the RL wrapped with MCTS (inspired by AlphaZero) agent for Connect-4. Therefore the AlphaZero-inspired approach performs the best, followed by a Minimax agent with Alpha-Beta cuts, then MCTS, and lastly, the classical RL agent.

Finally, in \cite{Dabas2022} it was shown that the best Connect-4 algorithm was MCTS, followed by Deep-Double Q-Learning then Minimax with Alpha-Beta pruning cuts. This was a surprising conclusion given that correctly configured Minimax algorithms play optimally \cite{Nasa2018, kang2019, Russell2021}. Furthermore, the Minimax payoff function is less sophisticated than existing literature sources (see \cite{Nasa2018, kang2019}) and the comparison between the algorithms was made on limited analysis suggesting incorrect conclusions.

It is hard to compare results from \cite{Dabas2022} and \cite{Scheiermann_2022} due to either the differences in algorithms formation (for example the payoff function in Minimax), or the differences in methodology (an algorithm versus algorithm approach in \cite{Dabas2022} as opposed to a base control algorithm as a comparison in \cite{Scheiermann_2022}). It is likely that in these differences lie the key to the variation in results. Another critical aspect was the limited evaluation. What if an algorithm plays perfectly but takes long time to make a move (see \cite{Escandon2018})? What if an algorithm plays strong against another but can not account for the variability of moves in a random agent (see \cite{Persson2011})?  

Motivated by these shortcomings, the contribution of this paper is threefold. First, we implement and compare advanced versions of the algorithms under review present in the literature to the case of Connect-4, enabling comparison with consistency over the methodology, branching factor and rules. Second, inspired by Axelrod's tournament \cite{Axelrod1981} for the Prisoner's Dilemma game, we introduce a novel methodology for comparative research, the Evolutionary Tournament. Third, we provide additional evaluation of each algorithm through a selection of classical evaluation approaches.

This paper is organized as follows. In Section~\ref{sec:methods}, we introduce each approach, their advanced counterpart and the evaluation methods. In Section~\ref{sec:results}, we present results and discuss the implications of each of the methods on the comparison. Finally, in Section~\ref{sec:conclusions}, we draw conclusions and present our future research directions.

\section{Algorithms}\label{sec:methods}
In this section, we introduce the design and implementation of the core algorithms used as candidates for the comparative analysis in Connect-4.

\textbf{Q-Learning}. Q-Learning is a temporal-difference (TD) algorithm where a Markov Decision Process (MDP), a framework for sequential decision problems \cite{Browne2012}, is considered. An MDP is defined as $MDP(\mathcal S, \mathcal A, P(s', s, a), R(s', s, a))$, where $\mathcal S$ is the set of possible states, $\mathcal A$ is the set of possible actions, $P(s', s, a) = P(s_{t+1} = s'|s_t = s, a_t = a)$ is the transition model that maps a new state $s'$ from state $s$ through action $a$ via one-step state transition dynamics, and $R(s', s, a) = \mathbf E [r_{t+1}|s_t = s, a_t = a,s_{t+1} = s']$ describes the reward function associated with the transition model obtained by the environment.

The goal in an MDP is to find a policy $\pi$ that maximises the expected reward over a given time horizon. This is described by equation (\ref{eq:bellman}), which is referred to as the Bellman equation:
\begin{equation}
V^\pi(s)=\sum_a \pi(s,a) \sum_{s'} P(s'|s,a)[R(s', s, a)+\gamma V^\pi(s')],
\label{eq:bellman}
\end{equation}
where the $V^\pi(s)$ is the value function that represents the expected utility under policy $\pi$ when starting in state $s$ and $\gamma \in [0, 1]$ is the discount factor, which ensures that the sum is bounded in an infinite horizon. %

Solving Equation \ref{eq:bellman} gives the optimal policy (mapping between the states and actions) \cite{Russell2021}. This is achieved by using the observed transitions to adjust the Q-Function values over time using the TD update equation:
\begin{equation}
Q(s,a) \leftarrow Q(s,a) + \alpha [R(s', s, a) + \gamma \max_{a} Q(s',a') - Q(s,a)],
\label{eq:TDupdate}
\end{equation}
where $\alpha \in [0,1]$ is the learning rate. In Q-Learning these values are recorded into a Q-Table for each state-action pair, $(s,a)$. A single Q-Table is used for our Q-Learning agents regardless of the position they play. During training each player will not have access to the alternative players Q-Table values as player I/II indexes the Q-Table based on states with even/odd number of moves. If a state-action pair is not visited then the respective utility value will not be stored in the Q-Table and instead represented as 0 or a missing index. Therefore, its vital that training has a balance between exploration and exploitation of the state space which an Epsilon-Greedy training policy promotes. Actions are selected at random with a probability equal $\epsilon$, a parameter, or selected according to a greedy policy \cite{Sutton2018} defined as $argmax_aQ(s, a)$ on the Q-Table \cite{Russell2021}, with probability equal to $1 - \epsilon$. This policy has been shown to be effective at training Q-Learning agents within Connect-4 contexts \cite{Wang2019}. The Q-learning algorithm with epsilon-greedy policy is provided in Algorithm 1.

\begin{table}[h]\normalsize
\begin{center}
\begin{tabular}{p{8cm}}\\
\toprule
\textbf{Algorithm 1} Q-Learning with Epsilon-Greedy \\
\midrule
\textbf{Input:} $\epsilon$, State $s$, Q-Table $Q(s, action)$, $\alpha$, $\gamma$\\
\textbf{Output:} $action$, updated Q-Table $Q'(s, action)$\\ 
$\quad 1:$ If $\epsilon$ $>$ randomValue: \\
$\quad 2: \quad$  $action$ = randomMove \\
$\quad 3:$ If $\epsilon$ $\le$ randomValue: \\
$\quad 4: \quad$  $action$ = $argmax_aQ(s, action)$ \\
$\quad 5:$ $s'$ = Function \textbf{UpdateState}($s$, $action$)\\
$\quad 6:$ $R$ = Function \textbf{Reward}($s'$, $s$, $action$)\\ 
$\quad 7:$ $FR$ = Function \textbf{FutureReward}($s'$, $action$):\\ 
$\quad 8: \quad$ $action' = argmax_aQ(s', action)$ \\
$\quad 9: \quad$ $s'$ = Function \textbf{UpdateState}($s'$, $action'$) \\
$\quad 10: \quad$\textbf{return} $MAX(Q(s', action'))$ \\
$\quad 11:$ $Q'(s, action)$ = Function \textbf{UpdateQ}($\alpha$, $\gamma$, $R$, $FR$)\\  
$\quad 12:$ \textbf{return} $Q'(s, action)$, $action$ \\
\midrule
\end{tabular}
\end{center}
\end{table}
Q-Learning has three hyperparameters, namely, a discount rate denoted by $\gamma$, a learning rate denoted by $\alpha$ and $\epsilon$. We set $\gamma = 0.5$ after testing a range of values against a random agent. Parameters $\epsilon$ and $\alpha$ are time-varying parameters that decay as a function of training iterations. This was shown to be effective in training Q-Learning agents in \cite{Wang2019} for epsilon decay and \cite{Russell2021, Persson2011, Arvidsson2010} for learning rate decay. The reward function, $R(s’, s, a)$, was designed in accordance with the Connect-4 literature \cite{Nasa2018, Dabas2022, kang2019}: Player win/lose (+40/-30), 1 piece in a row (+1), 2 pieces in a row (+2), 3 pieces in a row (+6), Column 2 and 6 placement (+1), Column 3 and 5 placement (+2), and Column 4 placement (+4). In the case where more than one condition is met, the values are summed.

We are now ready to introduce the first part of the first contribution of the paper, the comparison of base algorithms with enhancements. This paper tested whether the presence of an expert player during training improved the Q-Learning agents performance. As far as we are aware this is the first investigation of the type of enhancement applied to Connect-4. A Q-Learning agent was trained with half of its games played against a Minimax agent compared to another Q-Learning agent which played all of its games against itself. Minimax was selected as an expert player because it plays optimally if correctly configured \cite{Russell2021}. These two Q-Learning agents then played 100 games with the Q-Learning (Minimax) agent winning 61\% of its games. These results contradict those of \cite{Arvidsson2010} who found a Q-Learning agent was stronger when trained against a random player compared to more experienced advisories. A possible explanation for this is offered by \cite{Persson2011} which showed that surprising moves make a more challenging adversary. Therefore, the involvement of an expert player coaches the agent to respond to a better set of moves making the states experienced more varied and preparing the agent better. The Q-Learning (Minimax) agent was then re-trained for 100,000 games as player I and 100,000 games as player II totalling 200,000 games of experience, with the first 2000 games against a Minimax opponent and the rest through self-play. This constituted 1,293,018 unique Connect-4 states experienced across both player positions. Once the agent was trained, a Greedy policy was used for agent inference as this has been shown to be the optimal policy \cite{Sutton2018}.

\textbf{Minimax}. In a two-player game, the Minimax value, determined by equation (\ref{eq:MinimaxValue}), is the smallest payoff that the other player can guarantee the player will receive. The maximin value, determined by equation (\ref{eq:maxminValue}), is the largest payoff which the player can guarantee without knowing the other player's actions \cite{Maschler2013}. These are defined as:
\begin{equation}
    {\overline {v}}=\min_{a_I \in A_I}\max_{a_{II} \in A_{II}}{u(a_I,a_{II})},
\label{eq:MinimaxValue}
\end{equation}
\begin{equation}
   {\underline {v}}=\max_{a_I \in A_I}\min_{a_{II} \in A_{II}}{u(a_I,a_{II})},
\label{eq:maxminValue}
\end{equation}
where $A_I, A_{II}$ are the sets of strategies for player I and player II, respectively, and $u(\cdot)$ is the payoff function, representing the payment that player II makes to player I. For zero-sum two-player games such as Connect-4, maximising a players pay off is the same as minimising the opponent's payoff. Consider $U_I + U_{II} = 0$ where $U_I$ is utility which sums to zero representing one player with positive utility (winning) while the other with negative utility (a loss). Replacing each players individual utility with a shared utility, $U$, representing a payment player I makes to player II we get $U_I = U$ and $U_{II} = -U$ \cite{Maschler2013}. Therefore, $u(\cdot)$ captures the payoff of arriving at a state from player I's perspective. Equations \ref{eq:MinimaxValue} and \ref{eq:maxminValue} are equivalent to the Nash equilibrium of the game \cite{Maschler2013}.

Minimax search builds on these concepts by applying a recursive algorithm for selecting the best move in a 2-player (or n-player) game. This materialises as a plan from the perspective of two players (for Connect-4) which is captured in a sequential game tree structure detailing all possible moves from both players perspective. Each layer in the tree is ether player I or II. Due to the equations $U_I = U$ and $U_{II} = -U$, player I tries to maximise $u(\cdot)$ while player II tries to minimise $u(\cdot)$ \cite{Maschler2013}. In other words, the algorithm attempts to find the optimal move for each player given the assumption that the other player is playing optimally. 

A depth-first search is performed on the game tree beginning by considering the available actions at $S_0$ for the player. For example, if column 7 is full then the possible actions range from placing a piece in columns 1-6 from $S_0$. The algorithm chooses an action to get to a second state $S_1$ and then considers the list of available actions for the other player. Once an action has been selected, a new state is reached, $S_2$, where the algorithm once again considers the available actions for the player. This sequential process continues until a terminal state (win/draw/loss) or search \emph{depth}, set by $d$, is reached \cite{Russell2021}. For example, Minimax will search to 2-ply if $d = 2$ meaning in a sequential two player game the algorithm searches all game states where player I makes a move followed by player II making a return move. 

Once the sequential process ends, $u(\cdot)$ is called to evaluate the payoff in that game state with respect to player I. This payoff is feed backwards up the tree to the root node of the game tree, $S_0$. Dependent on the node type the payoff is fed to would result in differing values becoming populated in the tree. A player I/II node would favour higher/lower payoffs, calling equations (\ref{eq:maxminValue}) or (\ref{eq:MinimaxValue}). This is because each player chooses the action ($a_I$ or $a_{II}$) that rewards them, so player I/II would choose actions which maximise/minimise $u(\cdot)$. We backtrack to $S_0$ at which point the whole algorithm is repeated from the second action from the original six actions and again until all actions are explored. Once game tree is complete to a depth equal to $d$ the algorithm picks the action, $a_I$, which satisfies Equation \ref{eq:maxminValue} and represents maximised $u(\cdot)$ \cite{Russell2021}. The Minimax algorithm is provided in Algorithm 2.

\begin{table}[h]\normalsize
\begin{center}
\begin{tabular}{p{8cm}}\\
\toprule
\textbf{Algorithm 2} Minimax \\
\midrule
\textbf{Input:} State $s$, \emph{depth} $d$, $MaxBool$\\
\textbf{Output:} $action$\\
$\quad 1:$ If Function \textbf{IsTerminal}($s$):\\
$\quad 2: \quad$  \textbf{return} (Move, Function \textbf{ScoreState}($s$))\\

$\quad 3:$ ElseIf MaxBool == True:\\
$\quad 4: \quad$ $value$ = -$\infty$\\
$\quad 5: \quad$ For $action$ in availableActions:\\
$\quad 6: \quad \quad$ $s'$ = Function \textbf{UpdateState}($s$, $action$)\\
$\quad 7: \quad \quad$ $newScore$ = Function \textbf{Minimax}($s$,  $d$ - 1, 0)\\
$\quad 8: \quad \quad$ If $score$ $>$ $value$:\\
$\quad 9: \quad \quad \quad$ $value$, $move$  = $newScore$,  $action$\\
$\quad 10: \quad$\textbf{return} (move, score)\\

$\quad 11:$ ElseIf MaxBool == False:\\
$\quad 12: \quad$ $value$ = $\infty$\\
$\quad 13: \quad$ For $action$ in availableActions:\\
$\quad 14: \quad \quad$ $s'$ = Function \textbf{UpdateState}($s$, $action$)\\
$\quad 15: \quad \quad$ $newScore$ = Function \textbf{Minimax}($s$,  $d$ - 1, 1)\\
$\quad 16: \quad \quad$ If $score$ $\le$ $value$:\\
$\quad 17: \quad \quad \quad$ $value$, $move$  = $newScore$,  $action$\\
$\quad 18: \quad$\textbf{return} (move, score)\\
\midrule
\end{tabular}
\end{center}
\end{table}

Minimax has one parameter, $d$, which has a positive correlation with its strength but a negative correlation with run time. If depth is set to the maximum number of Connect-4 game moves ($d=42$) it would play a perfect game but would suffer significant increases in search time. A trade-off between time and optimally therefore exists \cite{Nasa2018} which we will evaluate as part of section \ref{sec:results}. The payoff function was designed according to Connect-4 literature \cite{Nasa2018, Dabas2022, kang2019}: win (+$\infty$/-$\infty$), 1 in a row (+1/-1), 2 in a row (+2/-2), 3 in a row (+6/-6), column 2 and 6 placement (+1/-1), column 3 and 5 placement (+2/-2), column 4 placement (+4/-4) where values are formatted as (Max/Min). In the case where more than one is activated, these values are summed.  

Continuing with our first contribution, Minimax was enhanced and compared to a variety of modifications beyond the base algorithm. Alpha-Beta Pruning has been shown to be a valuable modification for search efficiency \cite{Nasa2018} because large parts of the tree which do not affect the outcome are cut (i.e. pruned) by using acquired knowledge of explored sub-trees. This is moderated by two dynamic parameters, $\alpha$ and $\beta$, which initially are set to $\alpha = -\infty$ and $\beta = \infty$. As the search progresses, $\alpha$ and $\beta$ are updated to store the best values for player I/II in that particular sub-tree. If these values are found better than the current search then the search stops \cite{Russell2021}. Move ordering is another enhancement added to prioritize stronger moves earlier rather than delay a win. The success of algorithms such as Alpha-Beta depend on the order which nodes corresponding to actions are visited \cite{Russell2021} due to the depth first search. If the algorithm finds two win conditions, then both actions would be assigned large positive values and the first move visited would take priority over the second. Move Ordering aims to improve this by adding a small breadth search to the start of each node which provides an optimal sub-tree search order based on the next set of actions. This allows Minimax to perform more efficiently and search up to double the depth \cite{Russell2021}. Each combination of Minimax algorithms -- Minimax, Alpha-Beta (AB), Move Ordering (MO), and Alpha-Beta with Move Ordering (ABMO) -- played each other once in each player position (due to the deterministic nature of each) at Connect-4. Every game resulted in a draw. Minimax-ABMO was selected as the strongest Minimax variation due to evidence of higher efficiency \cite{Nasa2018, Russell2021} added by Alpha-Beta and Move Ordering. As far as we are aware, this is the first comparison between these enhancements and the base algorithm in the domain of Connect-4. 

\textbf{MCTS}. MCTS aims to solve the multi-arm bandit decision problem where an agent must choose between $A$ actions to maximise cumulative reward. MCTS does this by randomly sampling the decision space and then iteratively building a search tree outward. The algorithm, therefore, assumes that the true action value can be approximated by random simulation and that these approximations can be used to adjust the decision policy \cite{Browne2012}. Rather than use a reward function, MCTS estimates the action value using an \textit{average utility} of reward over several iterations of finished random games \cite{Russell2021}. There are four main stages which operate on the game tree in this order: selection, expansion, simulation, and backpropagation \cite{Russell2021}.

Selection requires a section policy to balance exploration and exploitation. \textit{Upper Confidence Bounds for Trees (UCT)} applies the Upper Confidence Bounds 1 (UCB1) selection policy \cite{Kocsis2006}:
\begin{equation}\label{eq:UCB1}
UCT = \frac{v_j}{n_j} + 2C_p \sqrt{\frac{2lnN}{n_j}},
\end{equation}
where $_j$ is node $j$ in the game tree; $v_j$, is the value of the outcome from the simulation phase; $n_j$, is the number of times that node $j$ has been visited by the algorithm; $N$ is the number of parent node visits; and $C_p$ is a parameter that controls the exploration limit \cite{Browne2012}. Equation \ref{eq:UCB1} calculates the upper confidence bound that a move is optimal \cite{Browne2012}. $\frac{v_j}{n_j}$ is the \textit{average utility} representing exploitation. As $\frac{v_j}{n_j}$ increases so does the UCT score for the node, meaning it is more likely to be selected. The right-hand side represents the exploration of the solution space. $C_p$ is called the exploration term with larger values representing the algorithms tendency to prefer exploration over exploitation \cite{Russell2021} and therefore equation \ref{eq:UCB1} addresses the exploration-exploitation dilemma. $C_p$ was selected using the theoretical value of $C_p = \frac{1}{\sqrt{2}}$ which has been shown to satisfy Hoeffding's Inequality \cite{Kocsis2006} representing an optimal choice for zero-sum games such as Connect-4 \cite{Browne2012}.

Starting at the root node, each child node is selected using (\ref{eq:UCB1}) with the child with the highest UCT value selected. If the child is a leaf node, then this is the final selection. If the child is not a leaf node, then its own child’s highest UCT value node is selected. This process is repeated until a leaf node is reached \cite{Russell2021}. If the leaf node has been visited by the algorithm before, then new children are added to the selected node in the game tree during the Expansion stage \cite{Russell2021}.

Simulation is applied from the selected leaf node if Expansion does not occur and applied to the first child of the leaf node if Expansion does occur \cite{Browne2012}. Simulation is a play-out of the entire game until a terminal state is reached, choosing moves for both player I and II with uniform distribution (random play). New nodes are not added to the game tree at this stage of the algorithm \cite{Russell2021}.

The outcome (win/lose as Connect-4 is zero-sum game) from Simulation is recorded for use in the Backpropagation stage. This stage updates $v_j$ and $n_j$ at each node from the simulated node all the way back to the root node \cite{Russell2021}. According to equation \ref{eq:UCB1}, when wins are backpropagated, $\frac{v_j}{n_j}$ increases meaning exploitation of the set of actions that lead to a win also increase and states that lead to a win/loss for the agent are more/less likely to be re-selected in the search process.

The above four steps are repeated until time, denoted $maxTime$, expires. At this point, the child node with the highest $\frac{v_j}{n_j}$ is selected and the move representing this state is used as a decision for the algorithm \cite{Russell2021}. $maxTime$ has a positive relationship with the strength of the decision \cite{Browne2012}. The MCTS algorithm is provided in Algorithm 3. 
\begin{table}[ht]\normalsize
\begin{center}
\begin{tabular}{p{8cm}}\\
\toprule
\textbf{Algorithm 3} Monte-Carlo Tree Search \\
\midrule
\textbf{Input:} State $s$, $maxTime$, $C_p$\\
\textbf{Output:} $action$\\
$\quad 1:$ $rootNode$ = Function \textbf{TreeNode}($n_j=0$, $v_j=0$)\\
$\quad 2:$ While $timeElapsed < maxTime$:\\
$\quad 3: \quad$ $node$ = $rootNode$\\
$\quad 4: \quad$ While $node$ is not leaf:\\
$\quad 5: \quad \quad$ $node$ = Function \textbf{Selection}($node$, $C_p$)\\
$\quad 6: \quad \quad$ $s$ = Function \textbf{UpdateState}($s$, $node$.move)\\
$\quad 7: \quad$ $node$.children = Function \textbf{Expansion}($s$)\\
$\quad 8: \quad$ While $s$ is not terminal:\\
$\quad 9: \quad \quad$ $s$ = Function \textbf{Simulation}($s$)\\
$\quad 10: \quad$ $result$ = Function \textbf{EvaluateWinLoss}($s$)\\
$\quad 11: \quad$ Function \textbf{Backpropagation}($node$):\\
$\quad 12: \quad \quad$ While $node$ has parent:\\
$\quad 13: \quad \quad \quad$ $v_j$, $n_j$ = $result$, $n_j$ + 1\\
$\quad 14: \quad \quad \quad$ $node$ = $node$.parent\\
$\quad 15:$ \textbf{return} $action$ = Function \textbf{BestMove}($rootNode$)\\
\midrule
\end{tabular}
\end{center}
\end{table}

Continuing with our first contribution of the paper, MCTS was enhanced and compared to the base MCTS algorithm with a modification called \textit{Decisive Moves} \cite{Teytaud2010} which alters the the stochastic nature of MCTS to avoid situations where good moves are overlooked \cite{Russell2021}. Decisive Moves ensure game deciding moves are controlled by a heuristic which ensures the algorithm takes any winning move it finds. If a winning move is not available but the opposing player could win in their next move, the heuristic ensures the oppositions move is blocked. If neither of these conditions are satisfied, then the UCT algorithm runs as normal. It is shown that this enhancement improves the strength of the algorithm with little additional computational costs when compared to the reference UCT method \cite{Teytaud2010}. In line with this, we found that the MCTS-Decisive algorithm won $60\%$ of games against the base MCTS algorithm showing the increased playing strength of the algorithm, an intuitive result because only the weaknesses, not the strengths, of MCTS are modified. As far as we are aware, this is the first comparison between the Decisive Moves enhancement and MCTS in the domain of Connect-4. 

\section{Evolutionary Tournament And Analysis} \label{sec:results}
The second and third contribution of this paper will be presented in this section. The second contribution is contained within the \textit{Evolutionary Tournament Evaluation} subsection and offers unique evolutionary process to analyse performance. The third contribution is an extensive analysis and is presented in the \textit{Algorithms Vs. Control Algorithms} and \textit{Relative Agent Evaluation} subsections. Previously, analysis has either concentrated on the relative performance of each algorithm when playing against each other \cite{Dabas2022}, or against common adversaries \cite{Scheiermann_2022}, or move times \cite{Nasa2018} making comparison difficult due to the varying conditions. We use all these approaches in one consistent setting thereby offering a wider level of analysis in a single place. For the remainder of this paper, an \textit{algorithm class} is defined as an algorithm type (Q-Learning, Minimax, MCTS) and each algorithm is denoted as the base class followed by enhancements and algorithm experience. For example, Minimax-ABMO-3 is Minimax with Alpha-Beta pruning, Move Ordering and $d$ = 3. Q-Learning (Minimax) will be simplified to Q-Learning.

\textbf{Evolutionary Tournament Evaluation}. Axelrod's tournaments \cite{Axelrod1981} were round-robin tournaments evaluating strategies for the Iterated Prisoner’s Dilemma. These tournaments helped provide researchers with better understanding of the successful strategies in understanding the evolution of cooperation \cite{Airiau2005} and serve as the inspiration for using evolutionary tournaments here.

\cite{Airiau2005} extends the tournament analysis by investigating the evolutionary mechanisms of a strategy demographic within a population to find the existence, nature, and convergence of the strategy distribution. We consider our agents akin to sub-species in a population allocating each agent a species count of ten. Simple evolutionary rules were implemented to simulate the evolutionary process over time. Two agents were randomly selected from the population (\textit{Tournament Selection} in \cite{Airiau2005}) to play a game of Connect-4. Each game of Connect-4 represents a generation with limited resources which two species fight over. If an agent wins then it reproduces (species count increases by one) to simulate domination over the other agent. The loser of the game dies and their species count decreases by one. While \cite{Airiau2005} uses scoring in their tournament, a 'death penalty' is suitable in Connect-4 because of the zero-sum nature of the game. In generation 1, players are initially selected with a uniform distribution due to the species population equality. Over time dominating species which win more games and reproduce giving them a higher chance of selection. This is called \textit{Selection Pressure} which drives the evolutionary process to favour stronger agents \cite{Miller1995}. This also acts as a robustness test because stronger algorithms are chosen more often and therefore must consistently win the games of Connect-4 against others.   

\begin{figure}[h!]
    \centering
    \includegraphics[scale = 0.36]{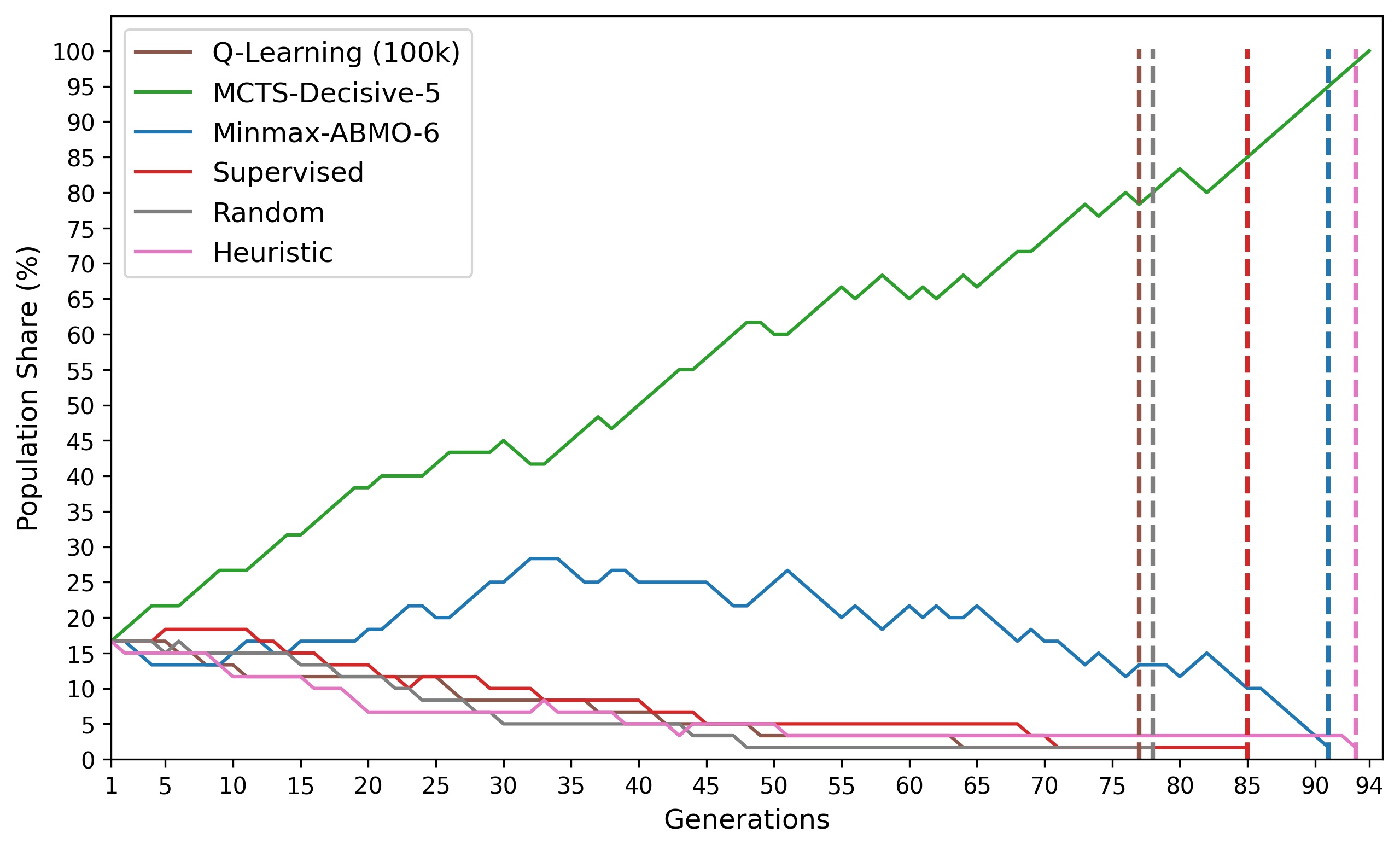}
    \caption[vscontrol]{Elite Evolutionary Tournament.}
    \label{fig:top_all}
\end{figure}

\begin{figure}[h!]
    \centering
    \includegraphics[scale = 0.36]{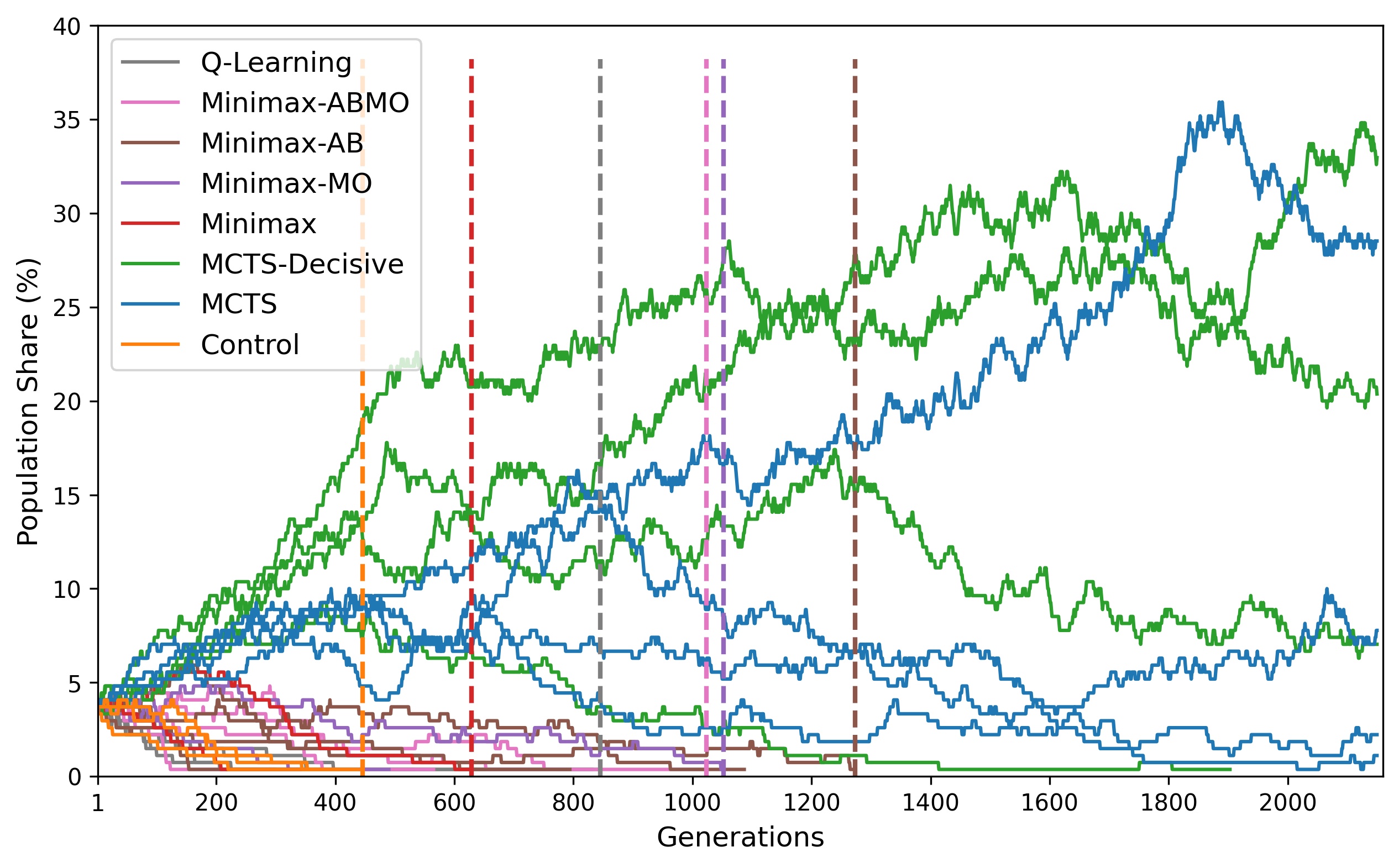}
    \caption[vscontrol]{Grand Evolutionary Tournament.}
    \label{fig:All_all}
\end{figure}

\begin{figure}[h!]
    \centering
    \includegraphics[scale = 0.36]{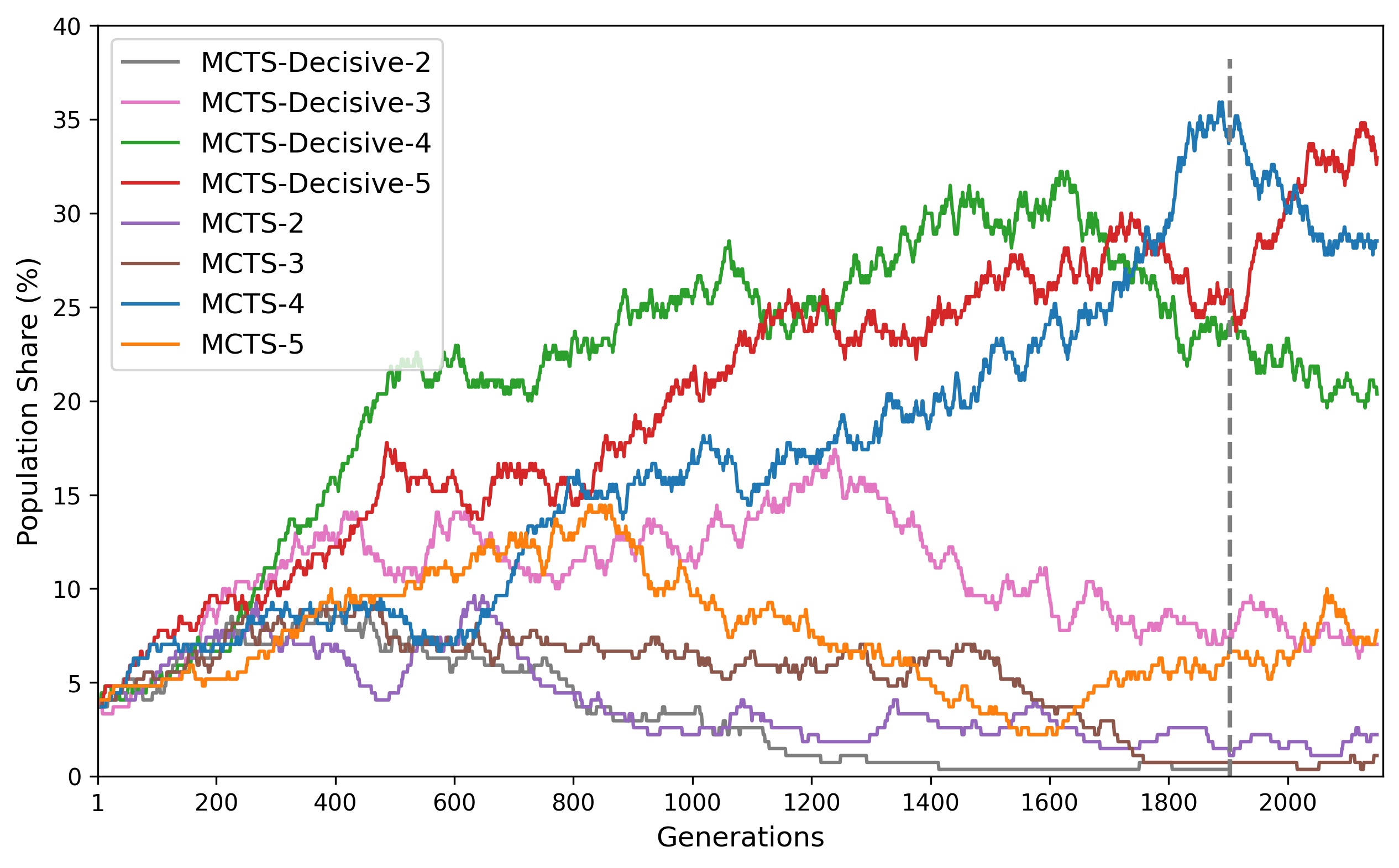}
    \caption[vscontrol]{Grand Evolutionary Tournament: MCTS Class.}
    \label{fig:All_mcts}
\end{figure}

Two tournaments are performed: an elite tournament (see Fig. \ref{fig:top_all}), consisting of just the enhanced variations and maximum experience levels in each class (MCTS-Decisive-5, Minimax-ABMO-6 and Q-Learning (100k), and a grand tournament (see Figs. \ref{fig:All_all}-\ref{fig:All_mcts}), containing 28 algorithms from every algorithm type and experience level. Experience was defined as games trained for Q-Learning, depth for Minimax, and seconds for MCTS and are described in section \ref{sec:methods}. Both tournaments included three standardised control algorithms. Evaluating against controls allows a standard metric to measure performance across the literature. Three controls were selected as opponents with their relative strength assessed in a multi-game round-robin style fixture. In ascending strength, these are: \textit{Random}, picks an action with a uniform probability distribution, see, e.g., \cite{Persson2011, Wang2019}; \textit{Supervised}, a semi-weak algorithm inspired by \cite{Schneider2002}; and \textit{Heuristic}, a rule-based algorithm which would choose the best move based on simple rules such as reward/payoff functions defined in section \ref{sec:methods}. 

Vertical dotted lines on each figure represent the generation that particular species was eliminated from the population. Flat solid lines represent no change in population share, positive solid lines represent increasing population share and negative solid lines represent declining population share. The elite tournament reached generation 94 while the grand tournament reached generation 2,151 before evidence of population share stability. 

In the elite tournament, MCTS-Decisive-5 was the strongest performing algorithm, eliminating all other species and finishing with a 100\% population share. A relatable equivalent of this behavior in evolutionary game theory is refereed to as an Evolutionarily Stable Strategy (ESS), which extends the concept of Nash equilibrium to a situation where the current population is immune to invasions \cite{Maschler2013}. This also shows the robustness of MCTS-Decisive-5 which consistently beat the other algorithms, such as the case after generation 83 where the trends of Minmax-ABMO-6 and MCTS-Decisive-5 were inversely associated. Between generation 76-80 and 82-94 we notice a steady population increase for MCTS-Decisive-5. Five significant events occur over these two periods: the elimination of the five other algorithm species. The selection pressure of MCTS-Decisive-5 increased as it started to dominate and it was repeatedly selected to play against the other algorithms. The second algorithm was Heuristic (eliminated at generation 93), third was Minimax-ABMO-6 (generation 91), fourth was Supervised (generation 85), fifth was Random (generation 77) and sixth was Q-Learning (100k) (generation 76). While Minimax-ABMO-6 was placed third, the results indicate that it was the second strongest algorithm because its population was largely was above its original share of 16\% and many of its upward trends correspond to the downward trends of other algorithms indicating its strength (for example, at generation 30 against MCTS-Decisive-5). Q-Learning (100k) was the worst performing algorithm getting eliminated before any of the controls. Based on previous evaluation we expected Q-Learning (100k) to outlast the random and possibly the supervised control but the stochastic nature of the selection mechanism contributed to this, for example generations 6-8 and 25-28 where Q-Learning (100k) was against the stronger opponents, MCTS-Decisive-5 and Minimax-ABMO-6 rather than weaker control algorithms.

The grand tournament showed the domination of the MCTS and MCTS-Decisive algorithm classes which all (except MCTS-Decisive-2) remained at the end of the tournament unlike the other 21 algorithms which were eliminated. The population of these classes began to stabilize after the last non-dominate class was eliminated and showed signs of fluctuation around a stationary point between generation 1,700-2,151 and therefore the tournament was stopped. Any small fluctuations are likely due to randomness in being selected as the first or second player where the former has the advantage of first move. When one algorithm does not capture 100\% of population share we can not definitively say that one algorithm was the strongest due to the stochastic nature of the selection. However, the evolutionary mechanism does show the domination of the MCTS type class clearly. The surviving algorithms were grouped into three sets around a population percentage representing relative strength of each group. Position one consisted of MCTS-4, MCTS-Decisive-4, and MCTS-Decisive-5; position two consisted of MCTS-5 and MCTS-Decisive-3; position three consisted of MCTS-2 and MCTS-3. The other algorithms positions were Minimax-AB (generation 1300), Minimax-MO (generation 1075), Minimax-ABMO (generation 1050), Q-Learning (generation 850), Minimax (generation 625), the controls (generation 450). MCTS-Decisive-2 was theoretically one of the weakest of the dominating MCTS type class' due to its low move time and therefore its elimination was not surprising. Minimax enhancements performed better individually (Minimax-AB, Minimax-MO) than as one algorithm (Minimax-ABMO). Q-Learning performed better than the Minimax base algorithm, similar to results in \cite{Dabas2022}. These results show a correlation between algorithm placement and move time and a clear positive impact of the Decisive Moves enhancement \cite{Teytaud2010}. 

\begin{figure}[h!]
    \centering
    \includegraphics[scale = 0.385]{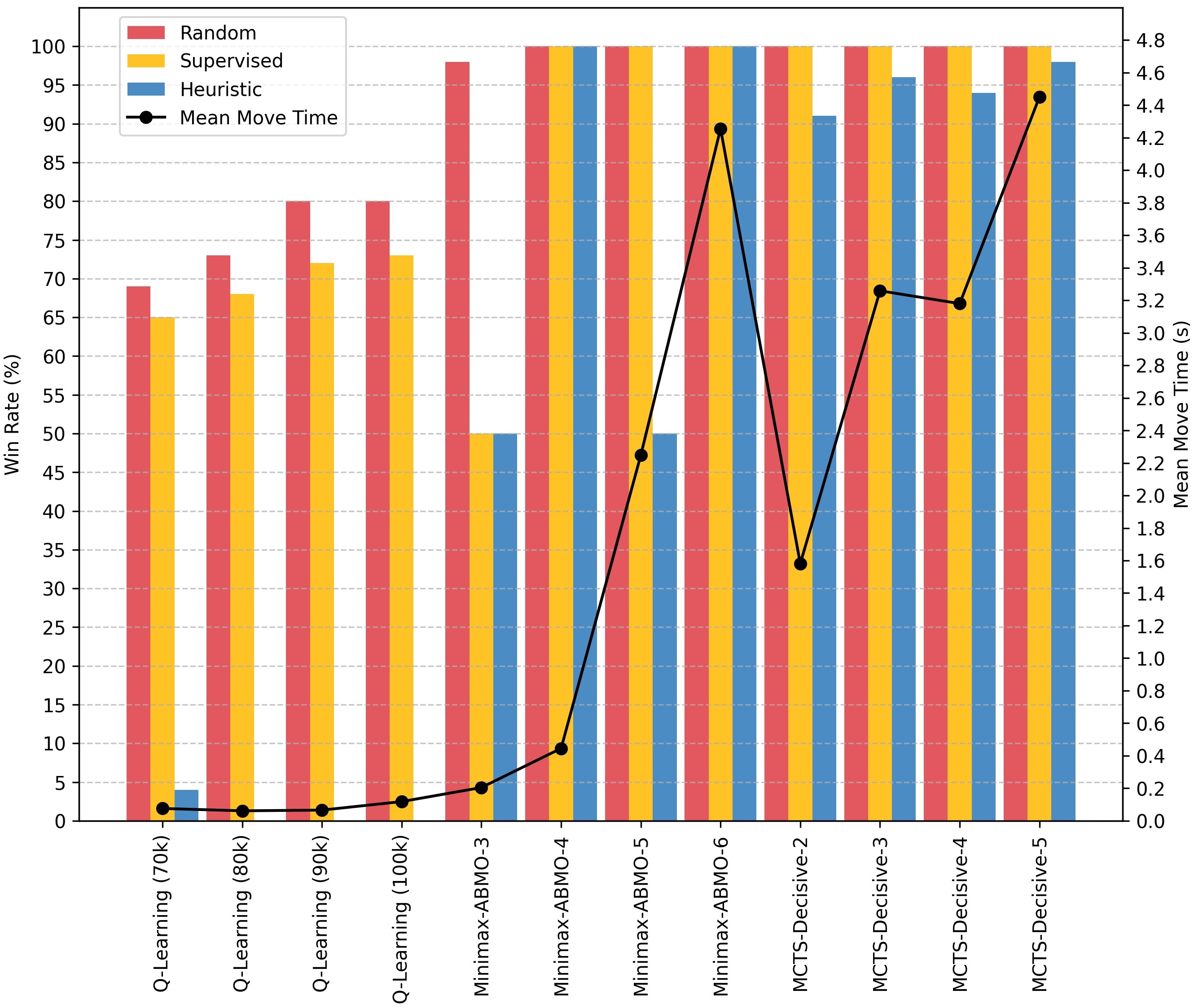}
    \caption[vscontrol]{Algorithm Vs. Control.}
    \label{fig:vscontrol}
\end{figure}

\textbf{Algorithms Vs. Control Algorithms}. The best performing agent in each algorithm class as defined in section \ref{sec:methods} was evaluated against three control algorithms, as defined in \textit{Evolutionary Tournament Evaluation}. Each algorithm class was assessed at differing experience levels against the control algorithms. Evaluation was conducted across two metrics: move timing, which is important for real time algorithm applicability \cite{Escandon2018}, and win rate (\%) \cite{Schneider2002, Persson2011, Wang2019}. Each algorithm-experience combination played 50 games against each control with a 50-50 split between player I and II to assess the strength of each algorithm with differing experience. Figure  \ref{fig:vscontrol} presents results on average move timing as well as win percentages for each of the algorithms against the control algorithms.
The Q-Learning algorithm class has the fastest mean move time with a high of 0.12s for Q-Learning (100k) and a low of 0.06s for Q-Learning (80k). Games trained and mean move time was positively correlated due to the increased size of the Q-Table as games experience increases. For Minimax-ABMO, the mean move time and depth are positively correlated with a high of 4.26s for Minimax-ABMO-5 and a low of 0.45s for Minimax-ABMO-4. This is in line with results reported in \cite{Nasa2018} where a Minimax Alpha-Beta agent $d = 4$ took 0.006s to make a move. MCTS mean move time and computation time are also positively correlated with a high of 4.45s for MCTS-Decisive-5 and a low of 1.58s for MCTS-Decisive-2. With the exception of MCTS-Decisive-3, the mean move time of each Decisive move enhanced algorithm is below its computation time, showing a more efficient outcome over the standard MCTS algorithm. This is because when the heuristic is activated it controls the move decision and not the MCTS process, resulting in a lower mean move time.

Q-Learning is the quickest class of algorithms with each variation taking less than 0.13s mean move time. Minimax was slightly better than MCTS in terms of speed with the upper limits of each algorithm class taking 4.26s and 4.45s and respectively. It is expected that Minimax would exponentially increase mean move time as the depth parameter increases \cite{Nasa2018} unlike MCTS as outlined in section \ref{sec:methods}. A adequate timing convention is currently lacking in the literature and therefore the importance of these results are defined by the application. Some may require quicker moves, while other applications could be more forgiving.

The win rates of all three algorithm classes increases with experience showing the learning success each algorithm at playing Connect-4. Against the Random control, all MCTS agents have a win rate of 100\%, Minimax has a mean win rate of 99.5\% due to Minimax-ABMO-3, and Q-Learning has a average win rate of 75.5\%. Against the Supervised control, all MCTS algorithm types have a win rate of 100\%, Minimax has a mean win rate of 87.5\% due to Minimax-ABMO-3 and Q-Learning has a mean win rate of 69.5\%. Minimax-ABMO-3 performs the worst out of all 12 algorithms against the Supervised control, indicating that a depth of 3 is too shallow for Minimax to play Connect-4 effectively. Against the Heuristic control, MCTS has a mean win rate of 94.75\%. Minimax has a mean win rate of 75\% due to the irregularity of performance across different depths:  $d=3$ and  $d=5$ (odd depths) win 50\% of games while $d = 4$ and $d=6$ (even depths) win 100\% of their games. This is contrary to the expectation that deeper searches produce a stronger performance \cite{Nasa2018} but are in line with results in \cite{Dabas2022} where a $d=2$ Minimax agent outperformed a $d=3$ agent against a constant advisory. The fact these results are repeated suggest that, for certain contexts, deeper does not always translate into a stronger performance. It could be possible that odd depths cause evaluation of the terminal state to become unbalanced as the opposing player moved last causing a lower max score compared to even depths. Similarly, \cite{Dabas2022} explains their results as the exploration of sub-optimal game trees caused by the odd search depth. Greater research over a larger range depths is needed for any solid conclusions. 

The Q-Learning agent performed poorly against the Heuristic agent with a mean win rate of just 1\%. This is likely explained because Q-Learning plays randomly in states it has not experienced yet \cite{Wang2019}. Due to the superiority of the Heuristic control compared to random play, its likely states not experienced were used highly and therefore Q-Learning under-performed compared to against other control algorithms.

\textbf{Relative Agent Evaluation}. Each algorithm (selected based on enhancement and experience from the previous evaluation) played a series of games against the others to assess their relative strengths. Rounds of 100 games between every combination of algorithms was played. These were split 50-50 between player I and II so each algorithm plays in the first and second position. Table \ref{t:algovsalgo} shows the mean win rate, draw rate, and move time of each combination of aggregated games.

\begin{table}[h]
    \centering
    \caption{Algorithms vs algorithms.}
    \label{t:algovsalgo}
    \begin{tabular}{|c|c|c|c|}
    \hline
        Algorithm & Vs. & Win & Mean Move \\
         ~ & ~ & Rate (\%) & Time (s)\\ \hline
        MCTS-Decisive-5 & Minmax-ABMO-6 & 69 & 5.69\\ \hline
        MCTS-Decisive-5 & Q-Learning (100k) & 100 & 2.12\\ \hline
        Minmax-ABMO-6 & MCTS-Decisive-5 & 27 & 3.4\\ \hline
        Minmax-ABMO-6 & Q-Learning (100k) & 100 & 4.25\\ \hline
        Q-Learning (100k) & MCTS-Decisive-5 & 0 & 0.11\\ \hline
        Q-Learning (100k) & Minmax-ABMO-6 & 0 & 0.05\\ \hline
    \end{tabular}
\end{table}

Q-Learning (100k) was the weakest in this evaluation losing every game against Minimax-ABMO-6 and MCTS-Decisive-5. On the other hand, the algorithm had the lowest mean move time, taking just 0.05s and 0.1s per mean move against Minimax-ABMO-6 and MCTS-Decisive-5 respectively. While this is encouraging for applications that requires a low mean move speed, the performance is so poor that it likely will not represent a viable solution to the game. Minimax-ABMO-6 was the second-best algorithm winning 100\% of games against Q-Learning (100k) and 27\% of its games against MCTS-Decisive-5. It took 3.4s mean move time against MCTS-Decisive-5 and 4.3s against Q-Learning (100k). The algorithms mean move time is lower than MCTS-Decisive-5 but higher than that of Q-Learning (100k) making the algorithm the middle ground for both performance and mean move time. Interestingly, the results show the algorithm is quicker against the stronger opponent (based on win rate). One possibility for this is that a tougher opponent would play a longer game and a weaker opponent would play a shorter game. Later game moves would require less calculations because there would be less remaining moves to search which would reduce mean move time. In contrast, early game moves would cause a higher mean move time as Minimax evaluates more possible moves to its search depth \cite{Russell2021}.

MCTS-Decisive-5 was the best performing algorithm winning 100\% of games against Q-Learning (100k) and 69\% of games against Minimax-ABMO-6. It took 5.7s mean move time against Minimax-ABMO-6 and 4.1s against Q-Learning (100k). This time difference is likely due to the relative strength of both the oppositions. Better moves would block more game deciding moves resulting in fewer activations of the Decisive Moves enhancement heuristic \cite{Teytaud2010} resulting in a slower mean move speed.

\section{Conclusion} \label{sec:conclusions}
In this paper, we have presented a comparative and systematic analysis of Q-learning, Minimax and MCTS. After formally introducing each of these algorithms, we have thoroughly compared their strengths and weaknesses on a single domain, namely, Connect-4. This has allowed us to directly assess the strengths and weaknesses of the algorithms in a common setting with identical branching factor. We have carried out this comparison through several approaches and introduced a novel method, the Evolutionary Tournament. Finally, we have showed that MCTS-Decisive is the strongest algorithm class in terms of win rate. This was supported by similar results in \cite{Dabas2022}. Q-Learning came first in terms of move speed, but its win performance was poor, likely due to a lack of training games compared to other literature implementations \cite{Arvidsson2010, Dabas2022}. Minimax held the middle ground in terms of both win rate and mean move speed. Future directions of work includes: i) enhancing the rules of the Evolutionary Tournament for comparative contexts enabling more accurate placement of non-dominant algorithms (See \cite{Airiau2005} for a score based approach), ii) expanding mathematical modeling of the tournament to evolutionary dynamics or discrete Markov chains to model the growth/decay of algorithm species, iii) investigating the impact of odd and even Minimax depths for contexts such as Connect-4 as found in our paper and in \cite{Dabas2022}.

\addtolength{\textheight}{-12cm}  




@book{Maschler2013, 
place={Cambridge}, 
title={Game Theory}, 
DOI={10.1017/CBO9780511794216}, 
publisher={Cambridge University Press}, 
author={Maschler, Michael and Solan, Eilon and Zamir, Shmuel}, 
year={2013}}

@Article{Airiau2005,
  author={Stephane Airiau and Sabyasachi Saha and Sandip Sen},
  title={{Evolutionary Tournament-Based Comparison of Learning and Non-Learning Algorithms for Iterated Games}},
  journal={Proceedings of the Eighteenth International Florida Artificial Intelligence},
  year=2005,
  pages={449-454},
   url={http://www.aaai.org/Library/FLAIRS/2005/flairs05-074.php}
}

@thesis{Arvidsson2010,
author  = {Oskar Arvidsson and Linus Wallgren},
title   = {Q-Learning for a Simple Board Game},
school  = {KTH Royal Institute of Technology, Stockholm, Sweden},
year    = {2010}, 
type={Bachelor’s Thesis}
}

@article{Browne2012,
title = {A survey of monte carlo tree search methods},
author = {Browne, {Cameron B} and Edward Powley and Daniel Whitehouse and Lucas, {Simon M} and Cowling, {Peter I} and Philipp Rohlfshagen and Stephen Tavener and Diego Perez and Spyridon Samothrakis and Simon Colton},
year = {2012},
volume = {4},
pages = {1-43},
journal = {IEEE Transactions on Computational Intelligence and AI in Games},
publisher = {IEEE},
number = {1}
}

@InProceedings{Dabas2022,
author={Dabas, Mayank
and Dahiya, Nishthavan
and Pushparaj, Pratish},
title={Solving Connect 4 Using Artificial Intelligence},
booktitle={International Conference on Innovative Computing and Communications},
year={2022},
publisher={Springer Singapore},
address={Singapore},
pages={727-735}
}

@INPROCEEDINGS{ Escandon2018,
author={Escandon, Elmer R. and Campion, Joseph}, 
booktitle={2018 IEEE XXV International Conference on Electronics, Electrical Engineering and Computing (INTERCON)}, 
title={Minimax Checkers Playing GUI: A Foundation for AI Applications}, 
year={2018}, 
pages={1-4}, 
doi={10.1109/INTERCON.2018.8526375}}

@article{kang2019,
  title={Research on Different Heuristics for Minimax Algorithm Insight from Connect-4 Game},
  author={Xiyu Kang and Yiqi Wang and Yanrui Hu},
  journal={Journal of Intelligent Learning Systems and Applications},
  year={2019},
pages={15-31},
volume = {11},
doi = {10.4236/jilsa.2019.112002}
}

@InProceedings{Kocsis2006,
author={Kocsis, Levente
and Szepesv{\'a}ri, Csaba},
title={Bandit Based Monte-Carlo Planning},
booktitle={Machine Learning: ECML 2006},
year={2006},
publisher={Springer Berlin Heidelberg},
address={Berlin, Heidelberg},
pages={282-293},
isbn={978-3-540-46056-5}
}

@article{Miller1995,
  title={Genetic Algorithms, Tournament Selection, and the Effects of Noise},
  author={Brad L. Miller and David E. Goldberg},
  journal={Complex Syst.},
  year={1995},
  volume={9}
}

@inproceedings{Nasa2018,
  title={Alpha-Beta Pruning in Mini-Max Algorithm –An Optimized Approach for a Connect-4 Game},
  author={Rijul Nasa and Rishabh Didwania and Shubhranil Maji and Vipul Kumar},
booktitle = {International Research Journal of Engineering and Technology (IRJET)},
  year={2018},
volume={5},
issue={4}
}

@thesis{Persson2011,
  author  = {John Persson and Tonie Jakobsson},
  title   = {Self-learning Game Player – Connect-4 with Q-learning},
  school  = {KTH Royal Institute of Technology, Stockholm, Sweden},
  year    = {2011}, 
type={Bachelor’s Thesis}
}

@book{Russell2021,
  title     = { Artificial intelligence: A Modern Approach, Global Edition},
  author    = { Stuart Russell and Peter Norvig},
  year      = {2021},
  publisher = { Pearson},
  address   = {Hoboken, New Jersey, USA},
edition = {4}
}

@article{Scheiermann_2022,
	doi = {10.1109/tg.2022.3206733},
	year = 2022,
	publisher = {Institute of Electrical and Electronics Engineers ({IEEE})},
	pages = {1--11},
	author = {Johannes Scheiermann and Wolfgang Konen},
	title = {{AlphaZero}-Inspired Game Learning: Faster Training by Using {MCTS} Only at Test Time},
	journal = {{IEEE} Transactions on Games}
}

@INPROCEEDINGS{ Schneider2002,
  author={Schneider, M.O. and Garcia Rosa, J.L.},
  booktitle={VII Brazilian Symposium on Neural Networks, 2002. SBRN 2002. Proceedings.}, 
  title={Neural Connect 4 - a connectionist approach to the game}, 
  year={2002},
  volume={},
  number={},
  pages={236-241},
  doi={10.1109/SBRN.2002.1181482}}

@article{Silver2016,
author = {Silver, David and Huang, Aja and Maddison, Chris J. and Guez, Arthur and Sifre, Laurent and van den Driessche, George and Schrittwieser, Julian and Antonoglou, Ioannis and Panneershelvam, Veda and Lanctot, Marc and Dieleman, Sander and Grewe, Dominik and Nham, John and Kalchbrenner, Nal and Sutskever, Ilya and Lillicrap, Timothy and Leach, Madeleine and Kavukcuoglu, Koray and Graepel, Thore and Hassabis, Demis},
  doi = {10.1038/nature16961},
  journal = {Nature},
  number = 7587,
  pages = {484--489},
  publisher = {Nature Publishing Group},
  title = {Mastering the Game of {Go} with Deep Neural Networks and Tree Search},
  volume = 529,
  year = 2016
}

@book{Sutton2018,
  title     = { Reinforcement Learning: An Introduction},
  author    = {Richard S. Sutton and Andrew G. Barto},
  year      = {2018},
  publisher = {MIT Press Ltd},
  address   = {Cambridge, Massachusetts, USA},
edition = {2}
}

@INPROCEEDINGS{Teytaud2010,
  author={Teytaud, Fabien and Teytaud, Olivier},
  booktitle={Proceedings of the 2010 IEEE Conference on Computational Intelligence and Games}, 
  title={On the huge benefit of decisive moves in Monte-Carlo Tree Search algorithms}, 
  year={2010},
  volume={},
  number={},
  pages={359-364},
  doi={10.1109/ITW.2010.5593334}}

@article{VANDENHERIK2002,
title = {Games solved: Now and in the future},
journal = {Artificial Intelligence},
volume = {134},
number = {1},
pages = {277-311},
year = {2002},
issn = {0004-3702},
doi = {https://doi.org/10.1016/S0004-3702(01)00152-7},
url = {https://www.sciencedirect.com/science/article/pii/S0004370201001527},
author = {H.Jaap {van den Herik} and Jos W.H.M. Uiterwijk and Jack {van Rijswijck}}
}

@article{Wang2019,
  author   = {Hui Wang and
               Michael Emmerich and
               Aske Plaat},
  title     = {Assessing the Potential of Classical Q-learning in General Game Playing},
  journal   = {Communications in Computer and Information Science},
  doi = {https://doi.org/10.1007/978-3-030-31978-6_11},
  year      = {2019},
  eprinttype = {arXiv},
  pages    = {138–150},
  volume = {1021}
}

@article{Axelrod1981,
author   = {R. Axelrod and W. D. Hamilton},
title     = {R. Axelrod and W. D. Hamilton, “The Evolution of Cooperation,” Science, vol. 211, no. 4489, pp. 1390–1396, 1981.}}
\end{document}